\documentclass{article}
\usepackage{ijcai23_numbered}
\usepackage{times}
\usepackage{soul}
\usepackage{url}
\usepackage[hidelinks]{hyperref}
\usepackage[utf8]{inputenc}
\usepackage[small]{caption}
\usepackage{graphicx}
\usepackage{amsmath}
\usepackage{amsthm}
\usepackage{booktabs}
\usepackage{algorithm}
\usepackage{algorithmic}
\usepackage[switch]{lineno}

\usepackage[utf8]{inputenc} %
\usepackage[T1]{fontenc}    %
\usepackage{hyperref}       %
\usepackage{booktabs}       %
\usepackage{amsfonts}       %
\usepackage{nicefrac}       %
\usepackage{microtype}      %
\usepackage{xcolor}         %

\usepackage{amssymb}
\usepackage[inline]{enumitem}
\usepackage[noabbrev,capitalize]{cleveref}
\usepackage{subcaption}  %

\usepackage{dblfloatfix}  %

\newcommand{\citep}[1]{\cite{#1}}
\newcommand{\citet}[1]{\citeauthor{#1}\shortcite{#1}}

\usepackage{makecell}    %
\usepackage{multirow}

\usepackage{xparse}                           %

\newcommand{\bb}[1]{\boldsymbol{#1}}  %
\newcommand{\set}[1]{\mathcal{#1}}

\renewcommand{\i}[1][]{\def\temp{#1}\ifx\temp\empty {} \else ^{(#1)} \fi}

\newcommand{\bx}[1][]{\bb{x}\i[#1]}
\newcommand{\br}[1][]{\hat{\bb{x}}\i[#1]}
\newcommand{\bz}[1][]{\bb{z}\i[#1]}
\newcommand{\by}[1][]{\bb{y}\i[#1]}

\newcommand{\btheta}{\boldsymbol{\theta}}
\newcommand{\bphi}{\boldsymbol{\phi}}
\newcommand{\beps}{\boldsymbol{\epsilon}}

\newcommand{\bmu}[1][]{\boldsymbol{\mu}\i[#1]}
\newcommand{\bsigma}[1][]{\boldsymbol{\sigma}\i[#1]}

\newcommand{\Y}{\mathbf{\mathcal{Y}}}  %
\newcommand{\X}{\mathbf{\mathcal{X}}}  %

\newcommand{\sY}{\mathrm{F}}  %
\newcommand{\sX}{\mathrm{N}}  %
\newcommand{\sZ}{\mathrm{D}}  %

\newcommand{\R}{\mathbf{\mathbb{R}}}

\newcommand{\Enc}{f_{\bphi}}
\newcommand{\Dec}{g_{\btheta}}

\newcommand{\qEnc}{q_{\bphi}}       %
\newcommand{\pDec}{p_{\btheta}}     %
\newcommand{\pTrue}{p_{\ast}}       %

\NewDocumentCommand{\argZX}{O{}O{}}{\def\temp{#1}\ifx\temp\empty {(\bz|\bx[#2])} \else {(z_#1|\bx[#2])} \fi}
\NewDocumentCommand{\argXZ}{O{}O{}}{\def\temp{#1}\ifx\temp\empty {(\bx|\bz[#2])} \else {(x_#1|\bz[#2])} \fi}

\NewDocumentCommand{\qzx}{O{}O{}}{\qEnc\argZX[#1][#2]}  %
\NewDocumentCommand{\pxz}{O{}O{}}{\pDec\argXZ[#1][#2]}  %
\NewDocumentCommand{\pz}{O{}}{\pDec(\bz)}        %
\newcommand{\txz}{\pTrue(\bx|\bz)}  %
\newcommand{\tz}{\pTrue(\bz)}       %

\newcommand{\Loss}{\mathcal{L}}

\newcommand{\LossReg}{\Loss_{\mathrm{reg}}}
\newcommand{\LossRec}{\Loss_{\mathrm{rec}}}
\newcommand{\LossVAE}{\Loss_{\mathrm{VAE}}}

\newcommand{\LossOverlap}{\Loss_{\mathrm{Overlap}}}

\newcommand{\braces   }[1]{\left\{#1\right\}}
\newcommand{\brackets }[1]{\left[#1\right]}
\newcommand{\parens   }[1]{\left(#1\right)}

\newcommand{\norm}[1]{\lVert #1 \rVert}

\newcommand{\E}{\mathbb{E}}

\newcommand{\Dkl}{D_{\mathrm{KL}}}
\newcommand{\Dsymkl}{\tilde{D}_{\mathrm{KL}}}

\newcommand{\Dist}{\mathrm{d}}

\newcommand{\VDistance}{\Dist_\mathrm{pcv}}  %
\newcommand{\GtDistance}{\Dist_\mathrm{gt}}

\title{Overlooked Implications of the Reconstruction Loss for VAE Disentanglement}

\author{
        Nathan Michlo\and
        Richard Klein\and
        Steven James
    \affiliations
        University of the Witwatersrand, Johannesburg, South Africa
    \emails
        nathan.michlo1@students.wits.ac.za,
        \{richard.klein, steven.james\}@wits.ac.za
}

\begin{document}
        \maketitle

\begin{abstract}

    Learning disentangled representations with variational autoencoders (VAEs) is often attributed to the regularisation component of the loss.
    In this work, we highlight the interaction between data and the reconstruction term of the loss as the main contributor to disentanglement in VAEs.
    We show that standard benchmark datasets have unintended correlations between their subjective ground-truth factors and perceived axes in the data according to typical VAE reconstruction losses.
    Our work exploits this relationship to provide a theory for what constitutes an adversarial dataset under a given reconstruction loss. 
    We verify this by constructing an example dataset that prevents disentanglement in state-of-the-art frameworks while maintaining human-intuitive ground-truth factors. Finally, we re-enable disentanglement by designing an example reconstruction loss that is once again able to perceive the ground-truth factors.
    Our findings demonstrate the subjective nature of disentanglement and the importance of considering the interaction between the ground-truth factors, data and notably, the reconstruction loss,
    which is under-recognised in the literature.

\end{abstract}

\section{Introduction}\label{sec:intro}
    
    A fundamental challenge in machine learning is discovering useful representations from high-dimensional data that can be used to solve subsequent tasks effectively.
    Recently, deep learning approaches have showcased the ability of neural networks to extract meaningful features from high-dimensional inputs for tasks such as classification \citep{krizhevsky2012imagenet} and reinforcement learning \citep{mnih2015human}.
    However, these learned representations are often not semantically meaningful, which can negatively impact interpretability, fairness~\citep{locatello2019fairness}, and downstream task performance~\citep{locatello2018challenging}.

    Prior work has therefore argued that it is desirable to learn a representation that is \textit{disentangled} \citep{bengio2013representation}.
    While there is no consensus on what constitutes a disentangled representation, it is generally agreed that it should be factorised so that each latent variable corresponds to a single explanatory variable responsible for generating the data \citep{burgess2018understanding}.
    For example, a single image from a video game may be represented by continuous latent variables governing the $x$ and $y$ positions of the player or enemies, as well as categorical variables governing their clothing or appearance.

    A common approach to discovering these representations is variational autoencoders (VAEs) \citep{kingma2013auto}, which are trained on unlabelled data to learn a lower-dimensional representation capable of reconstructing the input.
    However, it has been shown that unsupervised methods cannot reliably learn representations without the introduction of supervision or inductive biases \citep{locatello2018challenging}.
    The recently introduced Ada-GVAE framework partially overcame this problem by using a weakly supervised signal to discover underlying factors \citep{locatello2020weakly}, but there remains room for improvement.

    Interestingly, VAEs do not have an explicit mechanism that encourages the learning of disentangled representations, but it is theorised that this behaviour is related to the regularisation term and the information bottleneck principle \citep{burgess2018understanding,mathieu2019disentangling,rolinek2019variational}. 
    However, despite this hypothesis, there is still no explicit reason for why the representations learnt by these frameworks should align with generative factors in the data.
    Nonetheless, these frameworks have been shown to produce disentangled representations when trained on synthetically generated data, as measured by appropriate metrics \citep{eastwood2018framework,chen2018isolating,zaidi2020measuring}.

    In this paper,
    we aim to understand why VAEs implicitly learn disentangled representations by investigating the interaction between the reconstruction loss of the VAE and the input data.
    We provide compelling evidence that disentanglement occurs not because of special algorithmic choices or the regularisation term, but because of how VAEs perceive distances between observations in the datasets themselves according to the reconstruction loss, and the fact that these distances accidentally correlate to the chosen ground-truth factors generating the data.
    In particular, we find that \textit{standardised benchmarks are constructed in such a way that they unintentionally encourage models to learn what appear to be disentangled representations.}

    The main, summarised contributions%
    \footnote{Code is provided at: \url{https://github.com/nmichlo/disent}.}
    of this paper are:
    \begin{enumerate}[label=(\roman*)] \setlength{\itemsep}{0pt}
        \item We introduce the concept of \textit{perceived distance}, in terms of the VAE reconstruction loss, to measure overlap or similarity between dataset pairs. We demonstrate that perceived distances in existing datasets unintentionally correspond to the distances between ground-truth factors, and that VAEs learn these distances, explaining why learnt representations may appear disentangled.
        \item We provide a technique to visualise the correlation between perceived distances in the data and ground-truth factors generating the data. We use this understanding to provide a theory for what constitutes an adversarial dataset under a given reconstruction loss.
        \item We reveal the ineffectiveness of state-of-the-art models by using our theory to design a simple, example adversarial dataset with constant perceived distance between elements, over which VAE-based frameworks fail to learn disentangled representations.
        \item We provide an example solution to the adversarial dataset that modifies the reconstruction loss, and thus perceived distances across the dataset, so that VAE frameworks are again able to capture the ground-truth factors.
        \item We contribute
            \textit{Disent},
            a general PyTorch~\citep{paszke2017automatic} disentanglement framework, with common models, metrics, and datasets.%
            \footnote{
                Disent framework repository: \url{https://github.com/nmichlo/disent}.
                Code is provided under the MIT license.\label{footnote:code-url}
            }
    \end{enumerate}

\section{Background}

        Assume a dataset $\X = \braces{\bx[0], ..., \bx[n]}$ is a set of independent and identically distributed (i.i.d) observations $\bx \in \R^\sX$, generated by some random process involving an unobserved random variable $\bz \in \R^\sZ$ of lower dimensionality $\sZ \ll \sX$.
        Additionally, the true \textit{prior distribution} $\bz \sim \tz$ and true \textit{conditional distribution} $\bx \sim \txz$ are unknown. 
        Variational autoencoders (VAEs) aim to learn this generative process. 
        Unlike autoencoders (AEs), which consist of an encoder $\Enc(\bx) = \bz$ and decoder $\Dec(\bz) = \br$ with weights $\bphi$ and $\btheta$, VAEs instead construct a probabilistic encoder by using the output from the encoder or inference model to parameterise approximate posterior distributions $\bz~\sim~\qzx$. 
        The approximate posterior is then sampled during training to obtain representations $\bz$, which are then decoded using the generative model to obtain reconstructions $\br~\sim~\pxz$.
        
        A \textit{factorised Gaussian encoder} \citep{kingma2013auto} is commonly used. The posterior is modelled using a multivariate Gaussian distribution with diagonal covariance $\bz \sim \set{N}(\bmu_{\bphi}(\bx),\;\bsigma_{\bphi}(\bx))$, and the prior is given by the multivariate normal distribution $\pz = \set{N}(\mathbf{0},\;\mathbf{I})$, with a mean of $\mathbf{0}$ and diagonal covariance $\mathbf{I}$. 
        To enable backpropagation, the reparameterisation trick in \Cref{eq:reparameterization-trick} is used to sample from the posterior distribution by offsetting the distribution means by scaled noise values.%
            \footnote{
                The notation $\odot$ represents the element-wise product.
            }
        \begin{align}
            \bz &= \bmu_{\bphi}(\bx) + \bsigma_{\bphi}(\bx) \odot \beps,~\mathrm{where}~\beps \sim \set{N}(\mathbf{0}, \mathbf{I}) \label{eq:reparameterization-trick}
        \end{align}
        VAEs maximise the evidence lower bound (ELBO) by minimising the loss given by \Cref{eq:vae}. 
        VAE-based approaches often make slight modifications to this loss \citep{higgins2017beta,zhao2017infovae,hou2017deep,kumar2017variational,chen2018isolating,kim2018disentangling,locatello2020weakly}, but the terms of these modified loss functions can usually still be grouped into reconstruction and regularisation components, given by Equations \ref{eq:vae_recon} and \ref{eq:vae_regular} respectively. 
        The regularisation term affects the representations learnt by the encoder, while the reconstruction term improves the outputs from the decoder. 
        These terms usually contradict in practice, with strong regularisation leading to worse reconstructions but often better disentanglement \citep{higgins2017beta,burgess2018understanding}. %
        \begin{align}
            \LossRec(\bx, \br) &= \E_{\qzx} \brackets{\log \pxz} \label{eq:vae_recon}\\
            \LossReg(\bx) &= - \Dkl \parens{\qzx \;\|\; \pz} \label{eq:vae_regular}\\
            \LossVAE(\bx, \br) &= \LossRec(\bx, \br) + \LossReg(\bx) \label{eq:vae}
        \end{align}

    \subsection{Random Sampling Reorganises VAE Embeddings}

        Disentanglement in VAEs is generally attributed to the regularisation term in \Cref{eq:vae_regular}; however, we highlight that regularisation only enables the underlying disentanglement mechanism. Disentanglement arises rather as a result of VAEs reorganising the latent space to minimise reconstruction mistakes due to random sampling from the probabilistic encoder during training.
        Through this mechanism, a VAE will place similar observations according to the reconstruction loss in \Cref{eq:vae_recon} close together in the latent space \citep{burgess2018understanding,mathieu2019disentangling,zietlow2021demystifying}, as this action minimises sampling errors.

\begin{figure}
            \centering
            \includegraphics[width=1.0\linewidth]{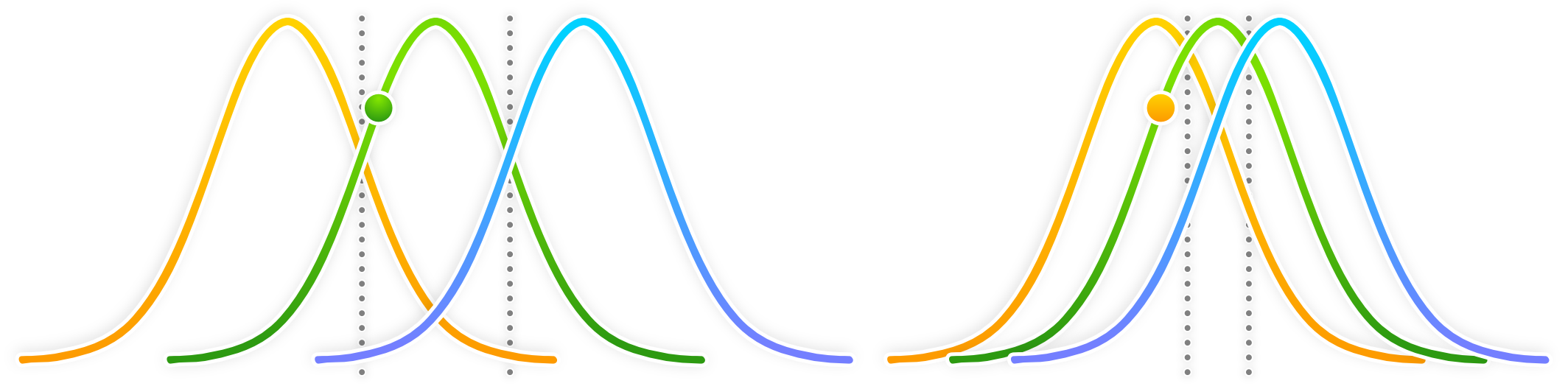}
            \caption{
                Nearby distributions in the latent space that correspond to different inputs. The VAE reconstructs a sample from the middle distributions. Left: weaker regularisation leads to few sampling mistakes, resembling a lookup table~\protect\citep{mathieu2019disentangling}. Right: strong regularisation leads to more reconstruction mistakes, where samples are attributed to nearby distributions, encouraging reorganisation.
            }
            \label{fig:random-sampling}
        \end{figure}

        The regularisation term enables this interaction by controlling the overlap between latent distributions corresponding to different inputs. If these distributions overlap sufficiently, the decoder will often attribute a random sample to an incorrect input, see \Cref{fig:random-sampling}. Thus, a mistake will be made during the decoding process, which encourages reorganisation to minimise the reconstruction error.

\section{Related Work}

    The following works are the most applicable to our research, falling under three general categories: 
    \begin{enumerate*}[label=(\roman*)]
    \item explanations for disentanglement,
    \item the role of the reconstruction loss in disentanglement, and
    \item problems with disentanglement.
    \end{enumerate*}

    Firstly, \citet{burgess2018understanding} relate VAEs to the information bottleneck principle. Which explains that random sampling leads to a local minimisation of the reconstruction loss which reorganises the latent space so that points close in pixel space are close in the latent space.
    \citet{mathieu2019disentangling} argue that VAEs do not explicitly encourage disentanglement through their design.
    Rather, they provide the explanation that the diagonal prior typically used in VAEs when combined with random sampling produces a similar effect to PCA.
    Our work takes inspiration from these ideas to develop the theory of perceived overlap in VAEs, which we use to analyse datasets and improve or hinder disentanglement.

    Secondly, inspired by \citet{burgess2018understanding} most modern frameworks offer some way to balance the regularisation and reconstruction components of the loss, the ControlVAE automates this process \citep{shao2020controlvae}.
    \citet{hou2017deep} instead swap out the reconstruction loss of VAEs for a perceptual loss function, which can improve the representations learnt by the model.
    \citet{zietlow2021demystifying} extend the analysis of \citet{mathieu2019disentangling};
    However, emphasis is placed on constructing adversarial datasets that hinder disentanglement performance using a mild transformation, obtained from trained models which achieve poor disentanglement scores.
    Our work provides intuition by constructing an example adversarial dataset that targets a specific reconstruction loss, and then remedies this problem by adjusting the loss.

    Finally, \citet{locatello2018challenging} show that useful representations cannot be reliably learnt with unsupervised methods, unless inductive biases are introduced, and
    \citet{gondal2019transfer} show that representations learnt on synthetic data often do not transfer well to real-world data.
    Our work investigates the interplay between the reconstruction loss and data as the main bias in VAEs, standard choices accidentally disentangle synthetic data.

\section{Existing Disentanglement Datasets}

    \begin{figure*}[b]
        \centering
            \begin{subfigure}{0.3038718436\linewidth}
                \includegraphics[width=1.0\linewidth]{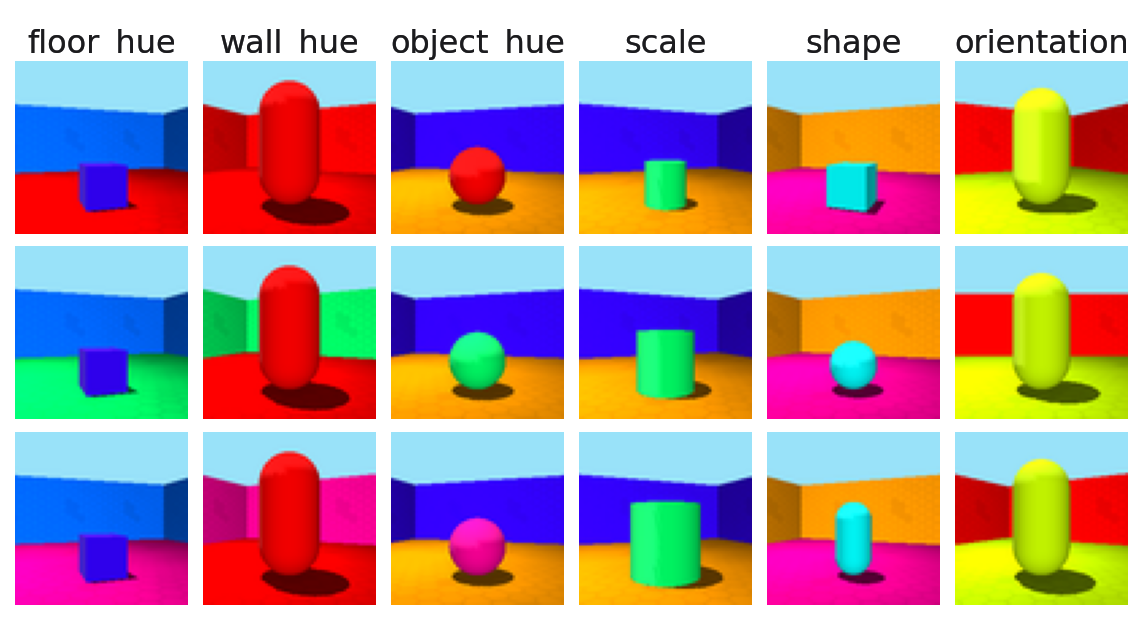}
                \caption{%
                    3D~Shapes
                }
                \label{fig:data-traversals_shapes3d}
            \end{subfigure}
        \hfill
            \begin{subfigure}{0.162579419\linewidth}
                \includegraphics[width=1.0\linewidth]{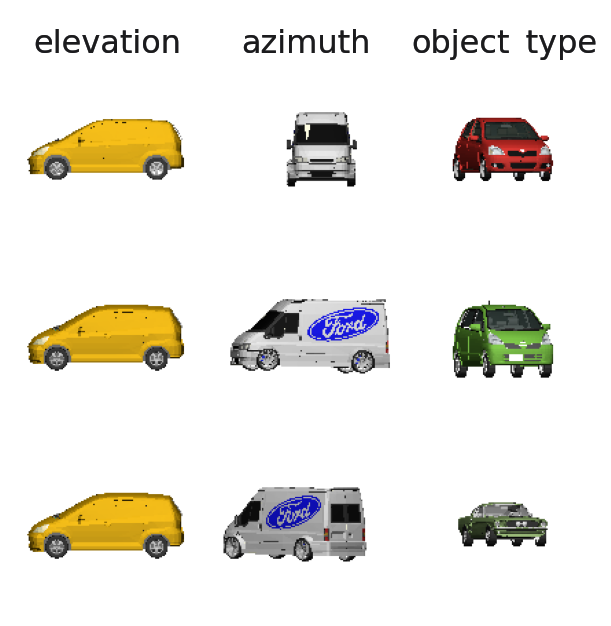}
                \caption{%
                    Cars3D
                }
            \end{subfigure}
        \hfill
            \begin{subfigure}{0.2567743687\linewidth}
                \includegraphics[width=1.0\linewidth]{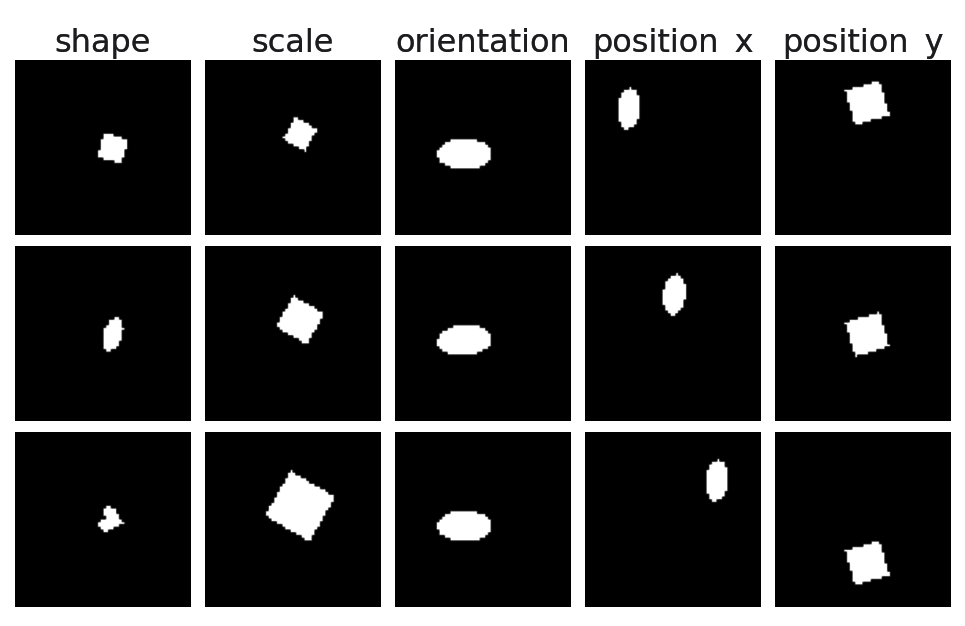}
                \caption{%
                    dSprites
                }
            \end{subfigure}
        \hfill
            \begin{subfigure}{0.2567743687\linewidth}
                \includegraphics[width=1.0\linewidth]{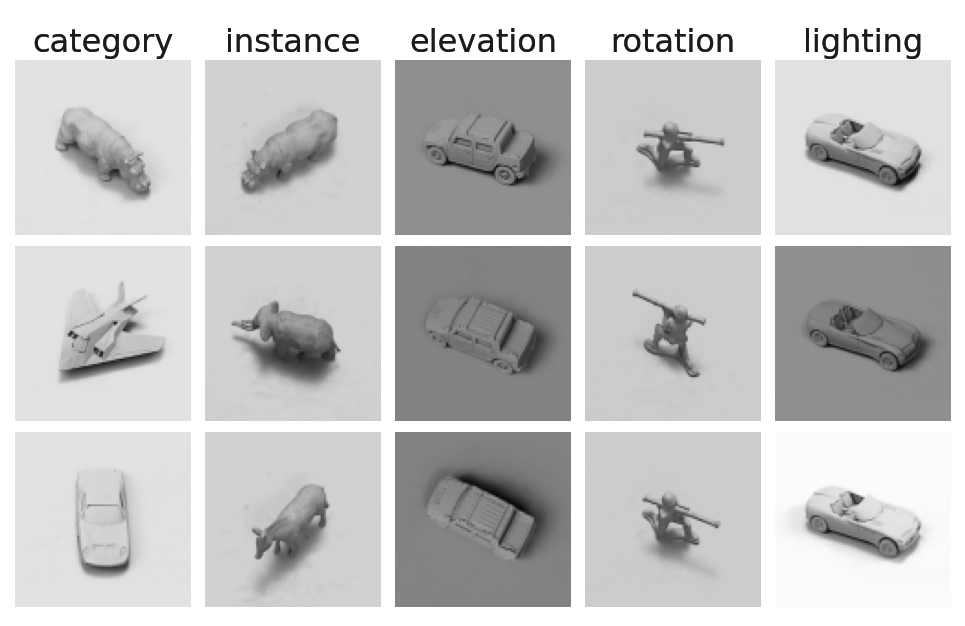}
                \caption{%
                    NORB
                }
            \end{subfigure}
        \caption{
            Common existing datasets used to benchmark disentanglement frameworks. These synthetic datasets are generated from ground-truth factors. Columns: %
            represent a traversal along a single factor.
        }
        \label{fig:data-traversals_shapes3d-cars3d}
        \label{fig:data-traversals_all}
    \end{figure*}

    Consider the 3D~Shapes dataset~\citep{3dshapes18} in \Cref{fig:data-traversals_shapes3d}, which contains observations of shapes fixed in the centre of the image with progressively changing attributes or factors such as size and colour. If, as humans, we are given unordered observations from a traversal along the size factor of 3D~Shapes, it would be easy to order these observations using a perceived increase or decrease in the size of the shape.
    We might even say that the shapes in the images overlap by different amounts, considering shapes that are closer in size to possess more overlap, and thus also considering them as closer together in terms of distance.
    This idea naturally extends to VAEs ordering pairs of observations, and so we seek to investigate the correspondence between how these frameworks perceive distances over data points to reorganise the latent space and the ground-truth factors themselves.

    \subsection{Dataset Ground-truth Distance} \label{section:define-gt-distance}

        Synthetic datasets \citep{3dshapes18,lecun2004learning,dsprites17,reed2015deep,gondal2019transfer} used for benchmarking disentanglement frameworks are all generated from $\sY \in \mathbb{N}^+$ ground-truth factors of variation. Each factor $i \in [\sY]$ represents some property about the data that can be varied,%
        \footnote{The bracket notation $[\sY]$ gives the natural numbers set $\{1, ..., \sY\}$.}
        and has a dimensionality or size of $f_i > 0$ where $f_i \in \mathbb{N}^+$.  The set of all factors used for generating the dataset is written as $\set{Y} = [f_1] \times ... \times [f_\sY]$. The full dataset is generated from this set of factors using some ground-truth generative process $\X = \braces{g_{\ast}(\by) \;|\; \by \in \Y}$. Examples of this generative process are given in \Cref{fig:data-traversals_shapes3d-cars3d}.

        With this construction of synthetic datasets, it is fitting to describe the \textit{ground-truth distances} between observations $\bx[a], \bx[b] \in \X$ using the Manhattan or $\ell_1$ distance between their corresponding ground-truth factors $\by[a], \by[b] \in \Y$, \footnote{
            When indexing $\by[a], \bz[a], \bx[a]$, for convenience $a$ may be an integer or a ground-truth factor $a = \by[a]$
        }
        as in \Cref{eq:gt-distance}. It is important to note that this choice may not be optimal for single factors; rather, $\ell_1$ distance naturally aligns with how the datasets are constructed. %
        \begin{align}
            \GtDistance(\bx[a], \bx[b]) &= \norm{\by[a] - \by[b]}_1. \label{eq:gt-distance}
        \end{align}

    \subsection{VAE Perceived Distance} \label{section:define-perceived-distance}

        With the idea of ground-truth distances between observations, we need a distance measure between observations as perceived by VAE frameworks. We derive the \textit{perceived distance} between dataset elements from the noisy sampling procedure and the chosen reconstruction loss in a VAE framework.

        Let $\bz[b] \sim \qzx[][a]$ be a (possibly incorrect) sample from the posterior distribution corresponding to some input element $\bx[a] \in \X$. Since the regularisation term encourages latent distributions to overlap, this sample $\bz[b]$ may be incorrectly attributed by the decoder to a distribution corresponding to some other element from the dataset $\bx[b] \in \X$, with reconstruction $\br[b] \approx \bx[b]$. 
        As the VAE objective consisting of the regularisation and reconstruction losses is jointly optimised, the decoder becomes better at reconstructing the inputs. In an ideal scenario, the inputs map to outputs ($\br \to \bx$), and reconstructions are samples from our dataset: $\br \in \X$. While this is not the case in practice due to the regularisation term, we derive the perceived distance in \Cref{eq:perceived-distance} from this assumption that $\br \to \bx$. This allows us to directly compare the elements $\bx[a], \bx[b] \in \X$ within a dataset using the reconstruction loss as a distance function:
        \begin{align}
            \VDistance(\bx[a], \bx[b])
            &= \lim_{\br \to \bx} \LossRec(\bx[a], \br[b])
            \\
            &= \LossRec(\bx[a], \bx[b]). \label{eq:perceived-distance}
        \end{align}
        
        The perceived distance depends on the choice of reconstruction loss, which in literature is usually the pixel-wise Mean Squared Error (MSE) for data that is assumed to be normally distributed. We assume MSE is used throughout the rest of this work, unless specified. Note that analyses are similar for other pixel-wise losses, such as Binary Cross-Entropy (BCE).

    \subsection{Perceived Distances Correspond to Ground-Truth} \label{section:distances-within-datasets}
    
        In \Cref{section:define-gt-distance}, all the ground-truth factors of a dataset are defined as the set $\set{Y} = [f_1] \times \ldots \times [f_\sY]$. In \Cref{eq:factor-traversal}, we now define a \textit{factor traversal} $\Y^{(a, i)} \subset \Y$ as the ordered set of all the coordinates along a factor $i \in [\sY]$ such that the set passes through a point $\by[a] \in \Y$. The number of elements in the traversal is equal to the size of the chosen factor $|\Y^{(a, i)}| = f_i$, and each element in the traversal generates the same traversal $\forall \by[b] \in \Y^{(a, i)}, \; \Y^{(a, i)} = \Y^{(b, i)}$. \Cref{fig:data-traversals_all} gives examples of traversals.
        \begin{align}
            \Y^{(a, i)}
            &= \; \ldots \times \braces{y^{(a)}_{i-1}} \times [f_i] \times \braces{y^{(a)}_{i+1}} \times \ldots
            \\
            &= \; \braces{(\ldots,\; y^{(a)}_{i-1},\; j,\; y^{(a)}_{i+1},\; \ldots) \;|\; \forall j \in [f_i] }
            \label{eq:factor-traversal}
        \end{align}

        We compute the distance matrix $\tilde{D}^{(a, i)} \in \R^{f_i \times f_i}$, for some distance function $\mathrm{d}$, between pairwise elements along a factor traversal $\Y^{(a, i)}$, written in \Cref{eq:distance-matrix} using matrix notation.
        \begin{align}
            \tilde{D}^{(a, i)} = \parens{\mathrm{d}(\bx[u], \bx[v])} \in \R^{f_i \times f_i} \; \forall u, v \in \Y^{(a, i)} \label{eq:distance-matrix}
        \end{align}

\begin{figure*}[t]
        \centering
            \begin{subfigure}{0.3094851794\linewidth}
                \includegraphics[width=1.0\linewidth]{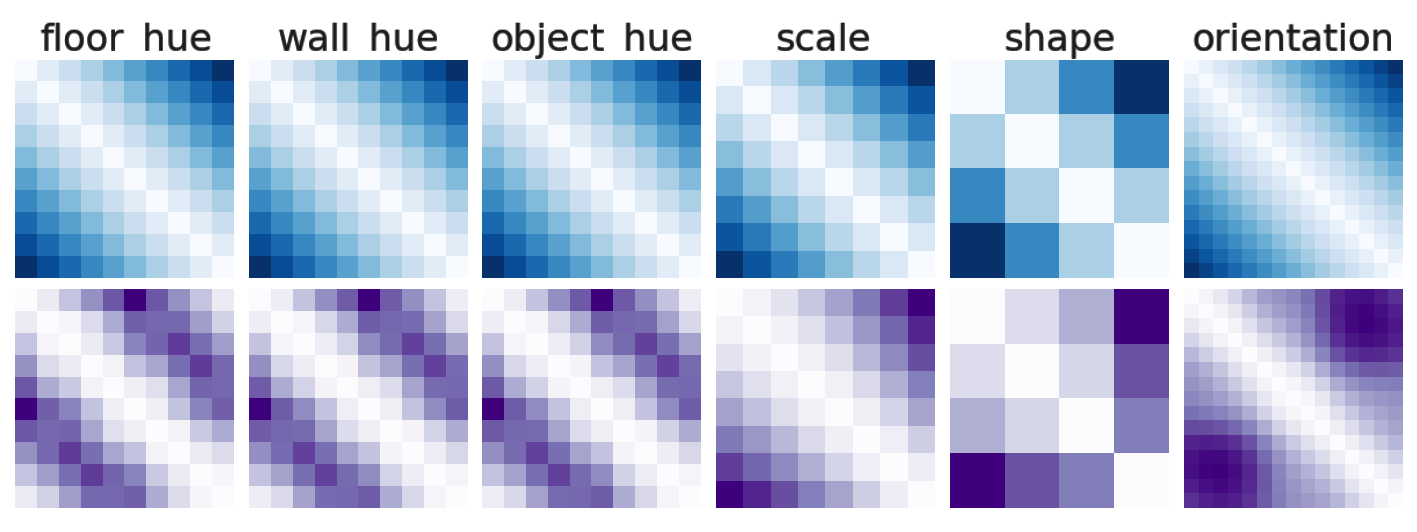}
                \caption{%
                    3D~Shapes
                }
            \end{subfigure}
        \hfill
            \begin{subfigure}{0.1546333853\linewidth}
                \includegraphics[width=1.0\linewidth]{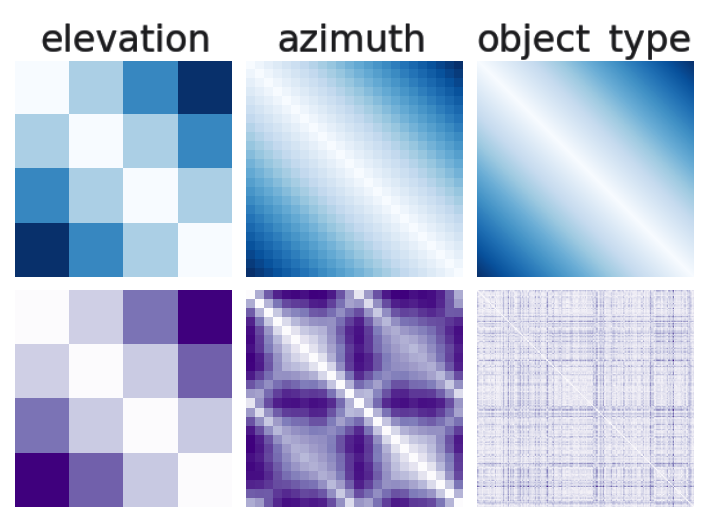}
                \caption{%
                    Cars3D
                }
            \end{subfigure}
        \hfill
            \begin{subfigure}{0.2579407176\linewidth}
                \includegraphics[width=1.0\linewidth]{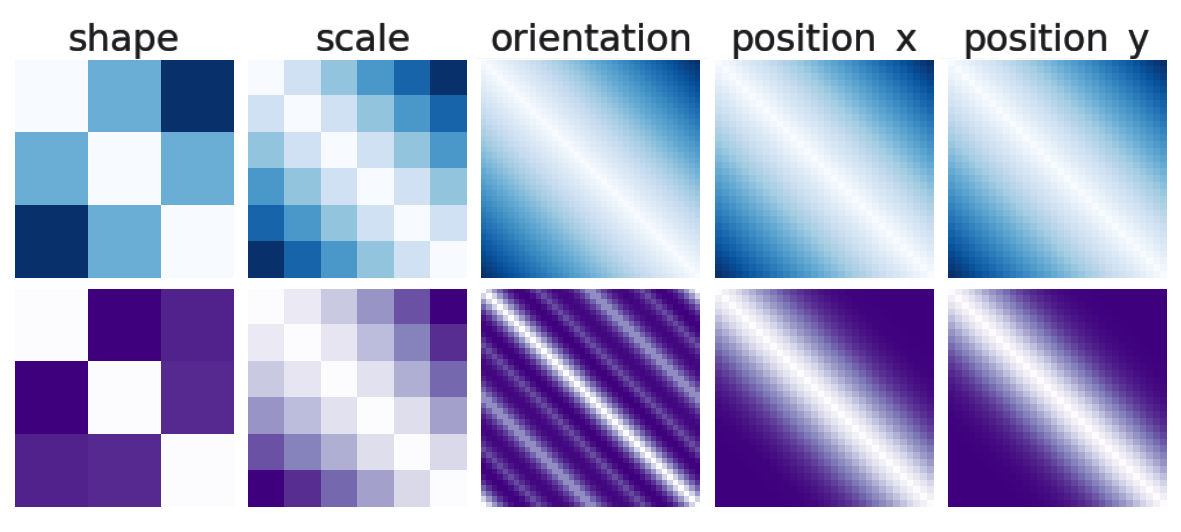}
                \caption{%
                    dSprites
                }
            \end{subfigure}
        \hfill
            \begin{subfigure}{0.2579407176\linewidth}
                \includegraphics[width=1.0\linewidth]{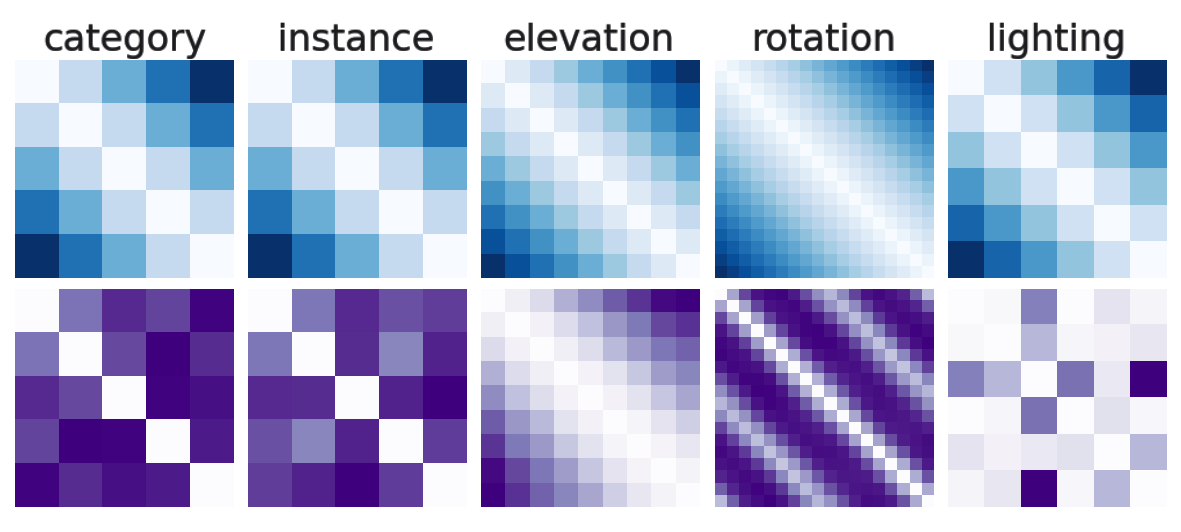}
                \caption{%
                    NORB
                }
            \end{subfigure}
        \caption{
            Distances in the ground-truth factor space naturally correspond to distances in the data space for current synthetic datasets. Top Row: Average ground-truth distance ($\ell_1$) matrices over factor traversals. Bottom Row: Average pixel-wise perceived distance (MSE) matrices over observations from the same factor traversals. Columns:
            Different ground-truth factors within each dataset.
        }
        \label{fig:data-dists_shapes3d-cars3d}
        \label{fig:data-dists_all}
    \end{figure*}

        To examine the ground-truth factors within our datasets, we compute the average distance matrix $D^{(i)} = \E_{a \in \Y}[\tilde{D}^{(a, i)}]$ for each factor $i \in [\sY]$. We plot these results in \Cref{fig:data-dists_shapes3d-cars3d} for both the ground-truth distance $\GtDistance$ and perceived distance $\VDistance$.
        It is immediately obvious from these plots that the ground-truth distances and the distances perceived by a VAE may accidentally correspond.

    Finally, we relate our work to \citet{burgess2018understanding} by outlining a direct approach in our extended paper in the supplementary material for computing relative factor importance using perceived distances.
    A factor is considered more important if a VAE prefers to learn it before another factor.

    \subsection{VAEs Learn Perceived Distances} \label{section:vaes-learn-perceived-distances}
    
\begin{figure}[h!]
            \centering
            \includegraphics[width=0.7\linewidth]{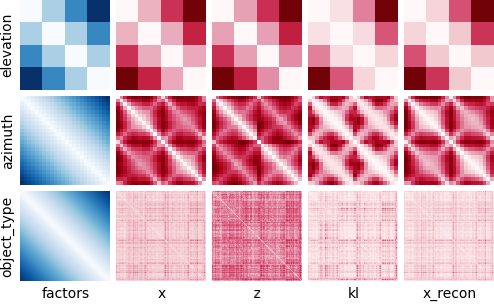}
            \caption{
                $\beta$-VAEs learn similar distances between observations at all levels of the network depending on the reconstruction loss. Rows: Different factors of the Cars3D dataset (Top to bottom: elevation, azimuth, car type), Columns: Distance matrices computed over factor traversals (Left to right: ground-truth distances, perceived distances between observations, $\ell_2$ distances over latent distribution means, KL divergences between latent distributions, perceived distances between reconstructions).
            }
            \label{fig:model-learns-dists}
        \end{figure}

        We compute distance matrices over a trained $\beta$-VAE at various levels of the network, including the representation layer and reconstructions. At each level of the VAE, the learnt distances all correspond to the original perceived distances already present within the dataset, see \Cref{fig:model-learns-dists}.
        Since VAEs reorganise the embedding space according to the perceived distances, and noting results from \Cref{section:distances-within-datasets}, VAEs may discover structures that are similar to the underlying ground-truth factors.

        Our results in \Cref{fig:data-dists_shapes3d-cars3d,fig:model-learns-dists} provide empirical evidence that VAEs mimic the distances already present in the dataset according to the reconstruction loss.
        To appear disentangled if the goal is factored representations, individual latent units will need to encode portions of the distances that correspond to factors within the data. However, it is known that VAEs with diagonal priors are rotationally invariant \citep{mathieu2019disentangling}, thus
        the same distances between the means $\bmu$ of latent distributions can be learnt for any arbitrary rotation of the latent space.
        This suggests that VAEs disentangle by accident, since ground-truth factors naturally correspond with distances in the dataset. If these perceived distances were to change such that they do not correspond to the ground-truth distances, VAEs might not be able to learn meaningful representations. This is highlighted by the fact that VAEs are already known to perform poorly on real-world data \citep{gondal2019transfer}.

\section{Adversarial Datasets} \label{section:adverserial-dataset}

    In the previous section, we highlighted the striking similarity between the ground-truth distances and the perceived distances between observations in synthetic ground-truth datasets.
    This suggests that disentanglement occurs because latent distances accidentally correspond to ground-truth distances, when the latent space is reorganised to minimise reconstruction errors and perceived distances from the data space are captured.

    Consider the example of a single chess piece moving across a chess board; there are no smooth transitions between grid points, since the piece is only valid when placed in the middle of squares.
    We describe such a dataset as having \textit{constant perceived distance}. This property is adversarial in nature as it is impossible for a VAE to order these observations using pixel-wise perceived distance.
    It is tempting to think that a harder case is if the perceived distances do not correspond to ground-truth distances; however, an (incorrect) ordering can then still be found. Existing datasets such as Cars3D already satisfy this incorrect ordering, which may explain the generally worse disentanglement performance compared to other datasets, see \Cref{fig:data-dists_all}.

    Formally, we say that a dataset has constant overlap when the pairwise distances over factor traversals are all equal. 
    Let $i \in [\sY]$ be a factor and $\by[a] \in \Y$ be ground-truth coordinate vector.
    Then, for all elements over the factor traversal $\forall \by[b] \in \Y^{(a, i)} / \braces{\by[a]}$, the corresponding perceived distance is constant such that $\VDistance(\bx[a], \bx[b]) = C_f$ with $C_f \in \R$ and $C_f > 0$. 
    Along factor traversals in such a dataset, no distinct ordering of elements can be found when a VAE tries to minimise the sampling error over the reconstruction loss. Going forward, we only consider the case where $\forall f \in [\sY], \; C_f = C$ for some $C > 0$.

    \subsection{Example XYSquares Adversarial Dataset} \label{section:xysquares-dataset}

\begin{figure}[bp]
            \centering
            \includegraphics[width=0.8\linewidth]{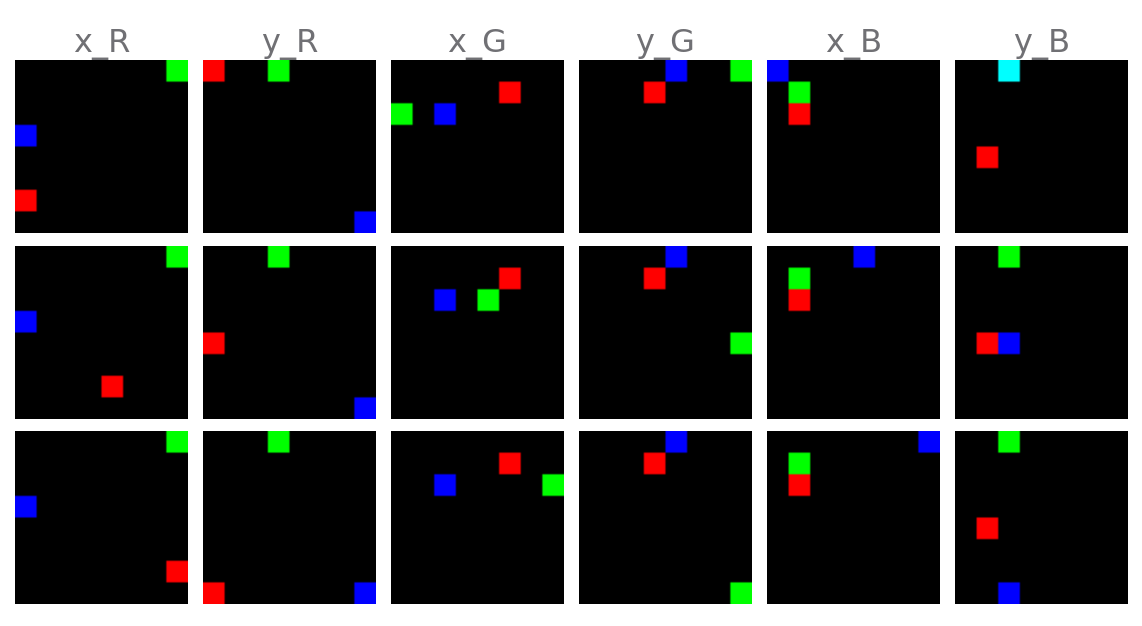}
            \caption{
               Columns represent ground-truth factor traversals over our adversarial XYSquares dataset. Pixel-wise losses measure constant values along these traversals.
            }
            \label{fig:data-traversal_xysquares}
        \end{figure}

        Taking inspiration from the chess piece example, we design a synthetic adversarial dataset called \textit{XYSquares} (See \Cref{fig:data-traversal_xysquares}) that specifically targets VAEs that use a pixel-wise reconstruction loss such as MSE, resulting in constant perceived distances.
        The dataset consists of three $8 \times 8$ pixel squares in a world of size of $64 \times 64$. This leaves $8$ grid positions along each axis without any pixel-wise overlap. The three squares are each assigned a colour according to R $(1, 0, 0)$, G $(0, 1, 0)$ and B $(0, 0, 1)$ to avoid any channel-wise overlap. With $6$ ground-truth factors (three squares moving along two axes), each with $8$ possible values, this gives a total dataset size of $8^6 = 262144$ observations.
        In the rightmost column of \Cref{fig:data-dists_xysquares-incr-overlap}, we validate that this leads to constant perceived distances between observation pairs in factor traversals.

    \subsection{Experimental Setup}

        We now investigate the performance of VAEs on our new dataset. 
        In particular, we use the unsupervised $\beta$-VAE \citep{higgins2017beta} and the state-of-the-art weakly supervised Ada-GVAE \citep{locatello2020weakly}. 
        The $\beta$-VAE scales the VAE regularisation term with a coefficient $\beta > 0$, while the Ada-GVAE breaks symmetry and encourages shared latent variables between pairs of observations. This is achieved by averaging together latent distributions between observation pairs that are estimated to remain unchanged when the KL divergence is below some threshold.
        We note that if the weakly supervised Ada-GVAE performs poorly, then it is highly likely that another unsupervised method will also perform poorly.

        We use the same Adam~\citep{kingma2014adam} optimiser and convolutional neural architecture as \citet{burgess2018understanding}. %
        To evaluate disentangled representations, we use the MIG~\citep{chen2018isolating} (Mutual Information Gap) and DCI Disentanglement~\citep{eastwood2018framework} scores. MIG measures the mutual information between the highest and second highest latent units for each factor, and DCI Disentanglement measures how much each latent unit captures a ground-truth factor using a predictive model.
        
        Finally, we perform an extensive hyper-parameter grid search for existing frameworks and datasets before running our own experiments.
        Hyperparameters include the learning rate, size of the latent dimension, training steps, batch size and $\beta$ values.
        See the supplementary material for further details on all experiments conducted throughout the remainder of the paper.

    \subsection{Example Adversarial Dataset Results}
    
        \begin{figure}[h!]
            \centering
            \includegraphics[width=1.0\linewidth]{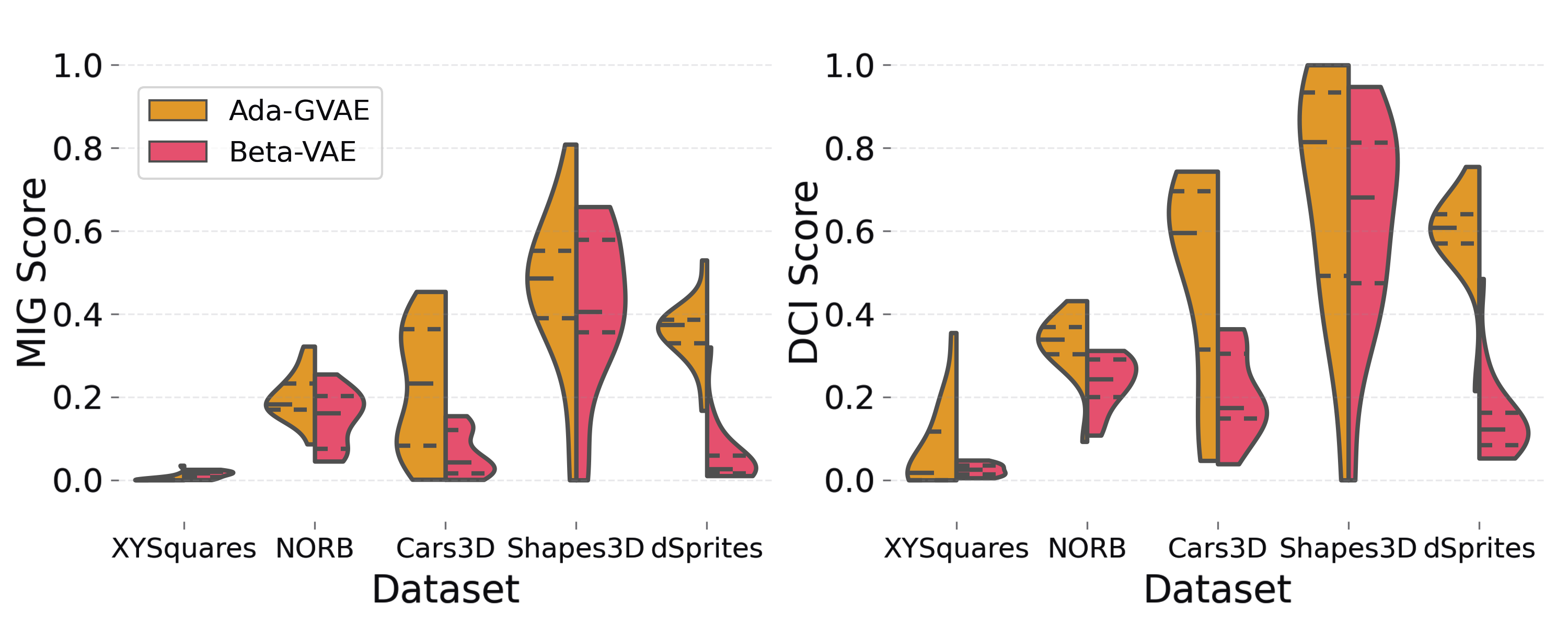}
            \caption{
                Densities over repeated runs for the attained MIG scores (left) and DCI Disentanglement scores (right) for the weakly-supervised Ada-GVAE (Left half of densities) and $\beta$-VAE (Right half of densities). XYSquares hurts the disentanglement performance significantly. Quartiles are marked with horizontal lines. We sweep over $\beta$ values and latent dimension sizes. See the supplementary material for details.
            }
            \label{fig:results-adverserial-dataset-vs-common}
        \end{figure}

\begin{figure}[h!]
            \centering
            \includegraphics[width=0.8\linewidth]{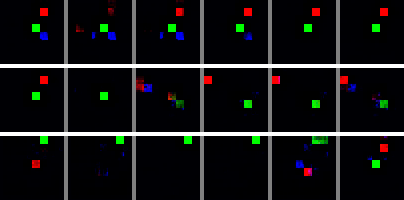}
            \caption{
               VAEs with pixel-wise losses fail to learn disentangled representations over the XYSquares dataset. Rows show latent traversals over a subset of latent units of a $\beta$-VAE. Varying one latent unit does not have an obvious effect or correspond to ground-truth factors.
            }
            \label{fig:latent-traversal_beta-vae_bad}
        \end{figure}

        \Cref{fig:results-adverserial-dataset-vs-common} shows that the disentanglement performance over XYSquares is extremely poor compared to existing datasets, even with the state-of-the-art Ada-GVAE.
        We are concerned only with the maximum score obtained for each model and dataset, as the graph is plotted over the hyper-parameter sweeps.
        This validates our adversarial dataset hypothesis in \Cref{section:adverserial-dataset}.
        Not only is the disentanglement performance poor, but much smaller values for $\beta$ are needed when tuning the regularisation loss.
        Example latent traversals from a VAE trained over the adversarial dataset are given in \Cref{fig:latent-traversal_beta-vae_bad}, results are far from disentangled and do not correspond in any way to the ground-truth factors in \Cref{fig:data-traversal_xysquares}.

    \subsection{Example of Varying Levels of Overlap}
        \label{section:varying-levels-of-overlap}
    
        We have examined the effect of training on existing datasets with significant amounts of overlap, as well as our own adversarial dataset with constant perceived distances according to pixel-wise losses.
        However, we have not investigated increasing levels of overlap in datasets, or rather reducing perceived between observations that are also close in ground-truth factor space.
        To do so, we modify XYSquares by decreasing the spacing between grid points while keeping the number of grid points constant along each factor, ensuring the dataset size remains fixed at $8^6 = 262144$ observations.

\begin{figure}[tp]
            \centering
                \includegraphics[width=1.0\linewidth]{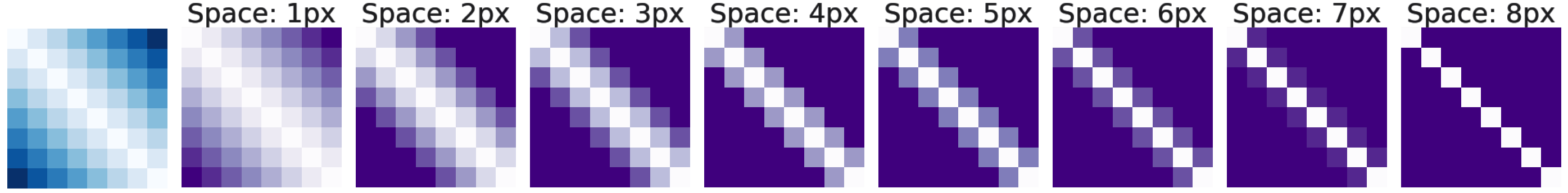}
            \caption{
                Ground-truth distance matrices (far left) and pixel-wise perceived distance matrices (left to right) over factor traversals. The spacing between grid-points of XYSquares decreases from 8px to 1px, which improves the correlation between perceived distances and ground-truth distances.
            }
            \label{fig:data-dists_xysquares-incr-overlap}
        \end{figure}

        The original adversarial dataset, with a spacing of $8$, has a constant distance value of $\VDistance(\bx[a], \bx[b]) = C_8$.
        As the spacing $s$ decreases from $8 \to 1$ over the datasets, the probability increases that any two observations re-sampled along a single factor traversal overlap $p(\VDistance(\bx[a], \bx[b]) < C_8)$ and should thus be placed closer together in the latent space. More overlap leads to more unique distance values which in turn allows for easier ordering of data points. We visualise this concept using ground-truth and perceived distance matrices in \Cref{fig:data-dists_xysquares-incr-overlap}.

\begin{figure}[tp]
            \centering
            \includegraphics[width=1.0\linewidth]{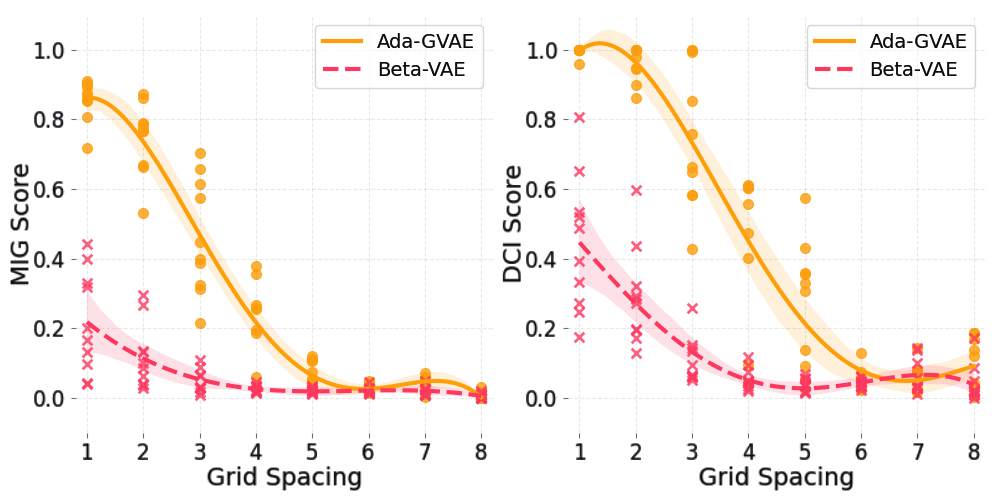} 
            \caption{
               XYSquares spacing vs disentanglment score (MIG -- left, DCI Disentanglement -- right). Decreasing (left to right) levels of overlap leads to decreased disentanglement performance. Each experiment is repeated 5 times with previously tuned hyper-parameters. See the supplementary material for further details.
            }
            \label{fig:exp-incr-overlap}
        \end{figure}

        We verify our statements through the experimental results in \Cref{fig:exp-incr-overlap}, where the $\beta$-VAE and Ada-GVAE are trained on these datasets. As the spacing decreases and overlap is introduced, the disentanglement performance improves, since it is easier for a VAE to introduce an ordering over representations. Even for the XYSquares dataset with 1 pixel of overlap between grid points, an ordering of elements along factor traversals can be induced. However, the probability of a VAE encountering these scenarios in the latent space due to random sampling is low, and thus it is still not always easy for the model to learn disentangled representations over such a dataset.
        
\section{Example of Introducing Overlap} \label{section:introducing-overlap}

    The previous section focused on increasing overlap by changing the underlying dataset; however, this still does not solve the case for the original XYSquares dataset with constant pixel-wise perceived distance.
    Throughout this paper, we have provided evidence that VAEs disentangle based on their reconstruction loss, which happens to align with ground-truth factors of variation in common benchmark datasets. This correspondence is not optimal for all tasks and we propose that this leads to the poor disentanglement performance in these settings. 
    Our solution is to choose a loss function that modifies perceived distances such that they also correspond to ground-truth distances.
    
    The new loss function we choose cannot be a pixel-wise approach, as this does not capture the distances due to the spatial nature of the XYSquares dataset.
    For the sake of simplicity in this example, we convert the existing pixel-wise loss function into a spatially aware loss function by introducing a differentiable augmentation to its inputs.
    An appropriate augmentation for our dataset is a channel-wise box blur.
    The problem, however, is that the decoder needs to be able to reconstruct the data, and so purely replacing the pixel-wise loss with the augmented loss may not succeed.
    Rather, in \Cref{eq:overlap-loss}, we append the augmented term to the existing loss and scale it by a constant $\alpha > 0$.
    \begin{align}
        \LossOverlap(\bx, \br) = \; \LossRec(\bx, \br) 
        + \alpha \, \LossRec(\mathrm{blur}(\bx),\, \mathrm{blur}(\br)) \label{eq:overlap-loss}
    \end{align}

    \subsection{Example Augmented Loss Experiments} \label{section:aug-loss-experiments}

\begin{figure}[tp]
            \centering
            \includegraphics[width=0.75\linewidth]{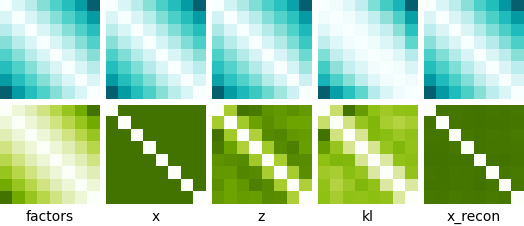}
            \caption{
                $\beta$-VAEs learn similar distances between observations. Top row: box blur augmented MSE. Bottom row: pixel-wise MSE loss. Columns: Distance matrices computed over factor traversals.
                All factors of XYSquares have the same statistics.
                This plot is constructed in a similar way to \Cref{fig:model-learns-dists}.
            }
            \label{fig:learnt-dists-mse-vs-overlap-loss}
        \end{figure}

        We choose a channel-wise box blur for our loss function with a radius of $31$, or a total kernel size of $63 \times 63$. We efficiently implement large filters using the Fast Fourier Transform. The size of the filter ensures that if two observations have active pixels on opposite sides of the images, then overlap will still be introduced between them. We set $\alpha = 63^2$, while this appears large, a box blur kernel is normalised so that the sum of all its values is $1$.
        We accordingly update our perceived distance measure and evaluate the new distances over the XYSquares dataset for each factor in \Cref{fig:learnt-dists-mse-vs-overlap-loss} after training and tuning $\beta$-VAEs.
    
\begin{figure}[bp]
            \centering
            \includegraphics[width=1.0\linewidth]{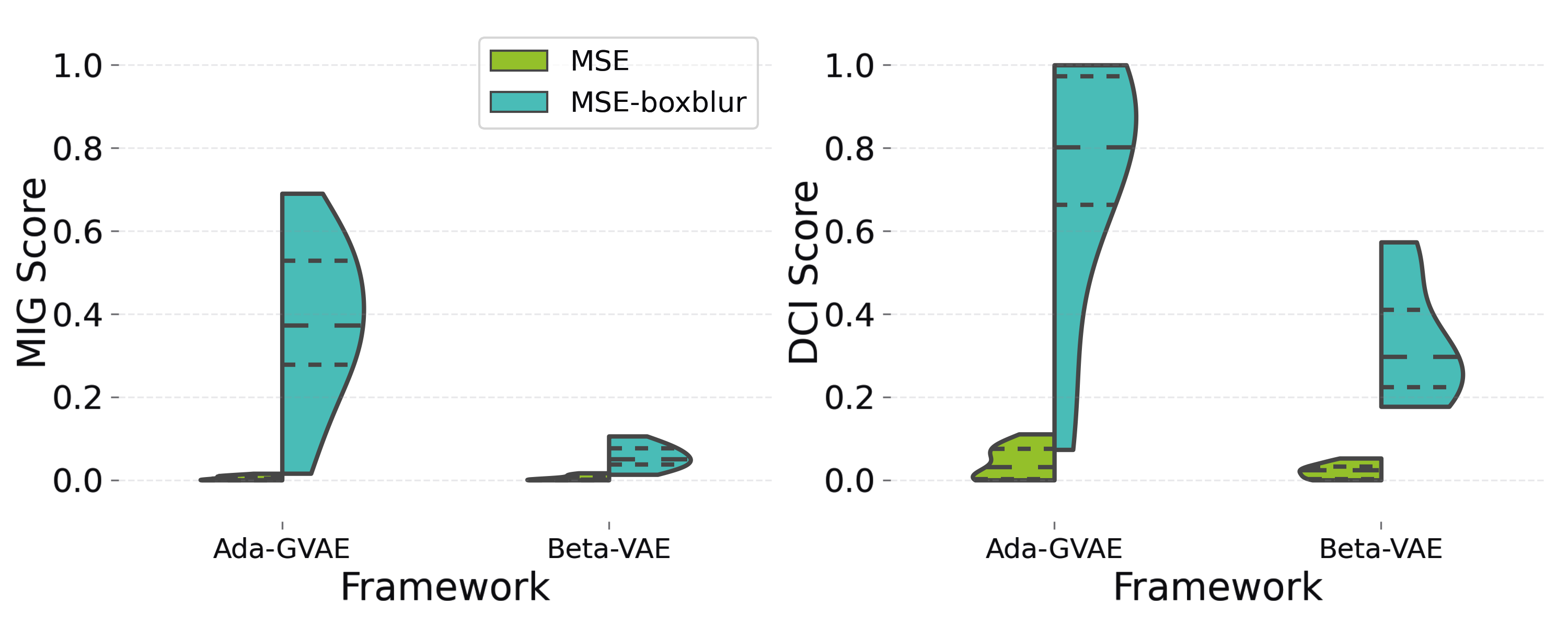}
            \caption{
                MIG and DCI scores for Ada-GVAE and $\beta$-VAE using the MSE loss and our modified loss function.
                Introducing a spatially aware loss function allows us to capture ground-truth distances between observations and allows the models to disentangle the adversarial XYSquares dataset.
            }
            \label{fig:results-overlap-loss}
        \end{figure}

        Finally, in \Cref{fig:results-overlap-loss}, we compare the performance of the spatially-aware loss function to the original pixel-wise loss. Our new loss significantly improves the disentanglement performance over the adversarial dataset. This is because it allows our models to capture perceived distances between observations that align with the ground-truth factors.
        
        While our choice of loss may not be optimal for disentanglement of these specific $x$ and $y$ factors from our adversarial dataset, disentanglement results are impressive. This is important because it provides the intuition that changing the loss function changes perceived distances and affects the ability of VAE frameworks to learn disentangled representations. 
        We leave learning or identification of optimal reconstruction losses for different datasets, to improve disentanglement, as future work.

 \section{Considerations for Disentanglement Research}

        We highlight the similarity between introducing overlap in \Cref{section:introducing-overlap} through the reconstruction loss function and varying levels of overlap in \Cref{section:varying-levels-of-overlap} through modifications to the construction of the dataset itself.
        Both methods aim to improve disentanglement by changing perceived distances to better correspond to the ground-truth factors, while keeping ground-truth factors fixed.

        The problem is that ground-truth factors can indeed change, and this choice, while at the discretion of the researcher, is largely ignored in literature. For example, a researcher may choose RGB, HSV or categorical representations for colours, they may choose binary or continuous encodings for positions, or they may split or merge various factors together.

        As our work shows, disentanglement is largely dependent on the chosen reconstruction loss and not special algorithmic choices. Obtaining improved disentanglement results under current VAE disentanglement frameworks will ultimately require supervision from the researcher to adjust perceived distances of the model to the task at hand.
        This contradicts the current notion that unsupervised and weakly supervised disentanglement methods can automatically uncover these human interpretable ground-truth factors \citep{higgins2017beta}.

        Ultimately, benchmarking against synthetic datasets with already subjective ground-truth factors will thus always remain problematic. There are infinitely many datasets with infinitely many choices as to what constitutes their ground-truth factors. Accurate disentanglement through future methods may need general world knowledge so that the methods can adapt to the task at hand.

\section{Conclusion}

        In this paper, we demonstrated that there are fundamental characteristics of existing datasets that encourage VAEs to learn disentangled representations.
        Our work provides a theory for how VAEs perceive distances between pairs of observations in datasets. We used this theory to provide intuition by constructing an adversarial dataset for pixel-wise losses over-which state-of-the-art VAEs fail to learn disentangled representations.
        Finally, we re-enabled disentanglement over the example adversarial dataset by again adjusting perceived distances, instead through a change of the VAE reconstruction loss to capture the ground-truth factors of the dataset.

        Our results highlight issues in current representation learning approaches.
        We find that the focus on regularisation for disentanglement is misplaced, rather, disentanglement is largely accidental, and careful choice of the reconstruction loss or data is needed to capture the ultimately subjective ground-truth factors.
        This is impractical in the real world, since perceived distances \textit{cannot} be a prerequisite for true disentanglement.
        More advanced methods are therefore required that can uncover true meaning within the data.

\section*{Acknowledgements}

Computations were performed using the High Performance Computing
infrastructure provided by the Mathematical Sciences Support unit
at the University of the Witwatersrand.

        \bibliographystyle{ijcai23_named}
        \bibliography{main}

        \clearpage
        \appendix
        \renewcommand{\thefigure}{A\arabic{figure}}
        \setcounter{figure}{0}
        \renewcommand{\thetable}{A\arabic{table}}
        \setcounter{table}{0}
        \twocolumn[
            \begin{@twocolumnfalse}
                \vspace*{0.5in}
                \begin{center}
                    {\LARGE\bfseries Overlooked Implications of the Reconstruction Loss for VAE Disentanglement\\[0.05in]Supplementary Material}
                \end{center}
                \vspace{0.6in}
            \end{@twocolumnfalse}
        ]
        
\label{footnote:extended-paper}

\section{Identifying Factor Importance} \label{appendix:identify-factor-importance}

    A factor in a dataset is considered more important if a VAE prefers to learn it before another factor. \citet{burgess2018understanding} identify this order of importance through a slow increase of the information capacity of VAEs during training. We note that simply by looking at the average perceived distance between observations along factor traversals, this ordering can be determined. Factors with a greater average distance will minimise the error in the reconstruction loss due to random sampling the most when learnt first. These factors (or components thereof) will thus generally be preferred.

    To compute the average perceived distance along a factor $f$, we sample a ground-truth coordinate vector $\by[a] \in \Y$ and then another random different coordinate vector $\by[b] \in \Y^{(a, i)}$ over the traversal for factor $f$ passing through $\by[a]$. Note that $\by[a] \neq \by[b]$. Then, we compute the perceived distance between the corresponding observations $\VDistance(\bx[a], \bx[b])$. We repeat this process to compute the expected perceived distance along factor traversals, given by \Cref{eq:ave-dist-factors}.
    \begin{equation}
        d_i = \E_{a \in \Y,\; b \in \Y^{(a, i)},\; a \neq b} \brackets{ \VDistance(\bx[a], \bx[b]) } \label{eq:ave-dist-factors}
    \end{equation}

     We determine the factor importance for dSprites as: $d_\mathrm{x}\approx0.058$ and $d_\mathrm{y}\approx0.057$ position, then $d_\mathrm{scale}\approx0.025$, then $d_\mathrm{shape}\approx0.022$, and finally $d_\mathrm{orientation}\approx0.017$. This aligns with the order determined by \citet{burgess2018understanding}. Computing estimates over an entire dataset can be intractable---for our estimates, we sample at least 50000 pairs per factor.

    Additionally, we compute the average perceived distance between any random pairs in the datasets (see  \Cref{eq:ave-dist-random}) and find that the average distance is higher. For dSprites specifically, we have $d_{\mathrm{ran}} \approx 0.075$. This suggests that the ground-truth factors correspond to axes in the data that minimise the reconstruction loss and is further evidence as to why VAEs appear to learn disentangled results.
    \begin{equation}
        d_{\mathrm{ran}} = \E_{a \in \Y,\; b \in \Y,\; a \neq b} \brackets{ \VDistance(\bx[a], \bx[b]) } \label{eq:ave-dist-random}
    \end{equation}

\subsection{Factor Importance Results}

    In \Cref{appendix:identify-factor-importance}, we relate our work to \citet{burgess2018understanding} by estimating the importance of different factors over the dSprites~\citep{dsprites17} dataset using the reconstruction loss (MSE) as the perceived distance function between observation pairs.

    We compute and list the order of importance of factors from the remaining datasets in \Cref{fig:factor-importance}. These importance values are computed as the average perceived distances between 50000 randomly sampled observation pairs taken along random factor traversals. Factors with higher average perceived distances will be prioritised by the model. For comparison, the average distance between any random pair in the dataset is also given. The average distances between pairs along factor traversals are usually less than the random distance, indicating that the ground-truth factors usually correspond to axes in the data that minimise errors.

    \begin{table}[h!]
        \centering
        \footnotesize
        \begin{tabular}{lrcc}
            \toprule
                \textbf{Dataset}
                & \textbf{Factor}
                & \textbf{Mean Dist.}
                & \textbf{Dist. Std.}
                \\
            \midrule
                 \multirow{4}{*}{
                    Cars3D%
                  } & \textbf{random} & 0.0519 & 0.0188 \\
                    & azimuth         & 0.0355 & 0.0185 \\
                    & object type     & 0.0349 & 0.0176 \\
                    & elevation       & 0.0174 & 0.0100 \\
                \hline

                \multirow{7}{*}{
                    3D~Shapes%
                } & \textbf{random} & 0.2432 & 0.0918 \\
                  & wall hue        & 0.1122 & 0.0661 \\
                  & floor hue       & 0.1086 & 0.0623 \\
                  & object hue      & 0.0416 & 0.0292 \\
                  & shape           & 0.0207 & 0.0161 \\
                  & scale           & 0.0182 & 0.0153 \\
                  & orientation     & 0.0116 & 0.0079 \\
                \hline

                \multirow{6}{*}{
                    Small~NORB%
                } & \textbf{random} & 0.0535 & 0.0529 \\
                  & lighting        & 0.0531 & 0.0563 \\
                  & category        & 0.0113 & 0.0066 \\
                  & rotation        & 0.0090 & 0.0071 \\
                  & instance        & 0.0068 & 0.0048 \\
                  & elevation       & 0.0034 & 0.0030 \\
                \hline

                \multirow{6}{*}{
                    dSprites%
                } & \textbf{random} & 0.0754 & 0.0289 \\
                  & position y      & 0.0584 & 0.0378 \\
                  & position x      & 0.0559 & 0.0363 \\
                  & scale           & 0.0250 & 0.0148 \\
                  & shape           & 0.0214 & 0.0095 \\
                  & orientation     & 0.0172 & 0.0106 \\
                \hline

                \multirow{3}{*}{
                    XYSquares
                } & \textbf{random} & 0.0308 & 0.0022 \\
                  & y (R, G, B)     & 0.0104 & 0.0000 \\
                  & x (R, G, B)     & 0.0104 & 0.0000 \\
            \bottomrule
        \end{tabular}
        \caption{
            Average perceived distances sampled along random factor traversals for different datasets. Components of factors with higher average distances will usually be prioritised by the model.
        }
        \label{fig:factor-importance}
    \end{table}
    
\begin{figure*}[h]
        \centering
        \includegraphics[width=0.7\linewidth]{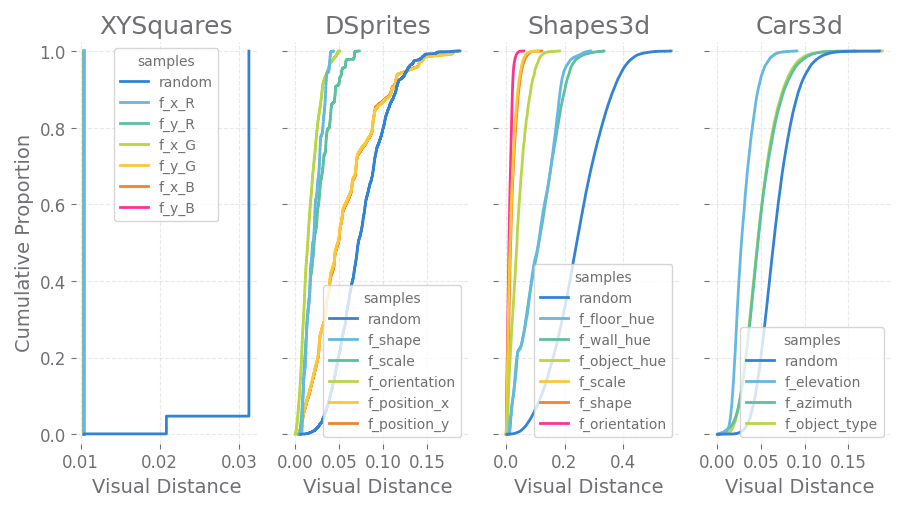}
        \caption{
            Cumulative proportion of perceived distance values between pairs sampled along factor traversals, compared to perceived distances between random pairs. Factors which are more important for the VAE to learn first to minimise the reconstruction loss have higher average perceived distances (lines shifted further to the right). This corresponds to the experimental results from \protect\citet{burgess2018understanding} which show that as the information capacity of a VAE is increased, it learns factors in order. For dSprites, this is $x$ and $y$ position, followed by scale, then shape, and finally orientation.
        }
        \label{fig:data-dists-freq}
    \end{figure*}

    We visualise the distribution of distances along factor traversals using cumulative frequency plots as in \Cref{fig:data-dists-freq}. It is interesting to note the distinct shift in structure for the adversarial XYSquares dataset, since distance values are constant depending on the number of differing factors.

\section{Implementation Details}

    In this section, we describe our various implementation details of the $\beta$-VAE~\citep{higgins2017beta} and Ada-GVAE~\citep{locatello2020weakly} frameworks, as well as the handling and standardisation of the different ground-truth datasets.

    \subsection{Beta Normalisation}
    
        For general consistency across datasets with different numbers of channels and models with different numbers of latent units, we implement beta normalisation as described by \citet{higgins2017beta}.

        Instead of taking the sum over the KL divergence in the regularisation term and the sum over elements in the reconstruction term of the VAE loss, we instead compute the means over elements in both terms and adjust the $\beta$ value accordingly.

    \subsection{Symmetric KL}
    
        The original Ada-GVAE implementation uses the asymmetric KL divergence $\Dkl \parens{p \;||\; q}$ as the distance function between the corresponding latent units of observation pairs. The Ada-GVAE uses this distance measure to estimate which of these latent distributions should be averaged together.
        
        We instead follow the approach of \citet{dittadi2020transfer} and use the symmetric KL divergence to compute these distances between latent units, improving the averaging procedure and computation of the threshold. The symmetric KL divergence is defined in \Cref{eq:symmetric-kl}.
        \begin{align}
            \Dsymkl \parens{p,\ q} = \frac{1}{2} \Dkl \parens{p \;||\; q} + \frac{1}{2} \Dkl \parens{q \;||\; p} \label{eq:symmetric-kl}
        \end{align}
        
    \subsection{Sampling Ada-GVAE Pairs} \label{appendix:sampling-adavae-pairs}
    
        The Ada-GVAE~\citep{locatello2020weakly} framework introduces weak supervision by sampling pairs of observations such that there are always $k \in [1, \sY]$ differing factors between them, where $\sY$ is the total number of factors generating the dataset. We use the weaker but more realistic case for sampling each pair, where $k$ is sampled uniform randomly from the range $[1, \sY]$ as described in the original paper.

    \subsection{Dataset Standardisation}
    
        For improved consistency and training performance, dataset observations are standardised. We first resize the observations to a width and height of $64 \times 64$ pixels using bilinear filtering if needed. Then the observations are normalised such that on average each channel of the image has a mean of 0 and a standard deviation of 1. Normalisation constants for each channel are precomputed across the entire dataset and are given in \Cref{fig:dataset-norm-constants}.
    
        \begin{table}[h!]
            \centering
            \footnotesize
            \begin{tabular}{lll}
                \toprule
                    \textbf{Dataset}
                    & \textbf{Mean}
                    & \textbf{Std}
                    \\
                \midrule
                    \makecell[l]{
                        Cars3D
                    }
                    & \makecell[l]{\textbf{R}: 0.897667614997663\\\textbf{G}: 0.889165802006751\\\textbf{B}: 0.885147515814868\\}
                    & \makecell[l]{0.225031955315030\\0.239946127898126\\0.247921063196844}
                    \vspace{0.75em} \\
                
                    \makecell[l]{
                        3D~Shapes
                    }
                    & \makecell[l]{\textbf{R}: 0.502584966788819\\\textbf{G}: 0.578759756608967\\\textbf{B}: 0.603449973185958\\}
                    & \makecell[l]{0.294081404355556\\0.344397908751721\\0.366168598152475}
                    \vspace{0.75em} \\
                
                    \makecell[l]{
                        Small~NORB
                    }
                    & \makecell[l]{0.752091840108860\\}
                    & \makecell[l]{0.095638790168273}
                    \vspace{0.75em} \\
                
                    \makecell[l]{
                        dSprites
                    }
                    & \makecell[l]{0.042494423521890\\}
                    & \makecell[l]{0.195166458806261}
                    \vspace{0.75em} \\
            
                    \makecell[l]{
                        XYSquares
                    }
                    & \makecell[l]{\textbf{R}: 0.015625\\\textbf{G}: 0.015625\\\textbf{B}: 0.015625}
                    & \makecell[l]{0.124034734589209\\0.124034734589209\\0.124034734589209}
                    \\
                \bottomrule
            \end{tabular}
            \caption{
                Precomputed channel-wise normalisation constants for datasets, assuming values of the input data are in the range $[0, 1]$.
            }
            \label{fig:dataset-norm-constants}
        \end{table}

\section{Experiment Hyper-Parameters}

    In this section, we give further details on the experiments conducted throughout the paper and their chosen hyper-parameters. For easier comparison with prior work, we use similar hyper-parameters, optimiser and model choices to \citet{higgins2017beta,kim2018disentangling,locatello2018challenging}.

    \subsection{Model Architecture}

        We use similar convolutional encoder and decoder models as \citet{higgins2017beta}. A full description of the basic VAE architecture is given in \Cref{fig:model-architecture}. The Gaussian encoder parameterises the mean and log variance of each latent distribution. The decoder uses the Gaussian derived Mean Squared Error (MSE) as the loss function. The number of input channels the encoder receives and the number of output channels the decoder produces depends on the dataset the model is trained on, this is either 1 or 3 channels.

        \begin{table}[h!]
            \centering
            \footnotesize
            \begin{tabular}{lll}
                \toprule
                    \textbf{Encoder} \\
                \midrule \\[0em]
                    \textbf{Input} & \{1 \textbf{or} 3\}x64x64 & \\
                    Conv.          & 32x4x4                    & (\textit{stride 2}, ReLU) \\
                    Conv.          & 32x4x4                    & (\textit{stride 2}, ReLU) \\
                    Conv.          & 64x4x4                    & (\textit{stride 2}, ReLU) \\
                    Conv.          & 64x4x4                    & (\textit{stride 2}, ReLU) \\
                    Linear         & 256                       & (ReLU)                    \\
                    2x Linear      & \{9 \textbf{or} 25\}      & \\
                    \\
                \toprule
                    \textbf{Decoder} \\
                \midrule \\[0em]
                    \textbf{Input} & \{9 \textbf{or} 25\}    &  \\
                    Linear         & 256                     & (ReLU) \\
                    Linear         & 1024                    & (\textit{reshape 64x4x4}, ReLU) \\
                    Upconv.        & 64x4x4                  & (\textit{stride 2}, ReLU) \\
                    Upconv.        & 32x4x4                  & (\textit{stride 2}, ReLU) \\
                    Upconv.        & 32x4x4                  & (\textit{stride 2}, ReLU) \\
                    Upconv.        & \{1 \textbf{or} 3\}x4x4 & (\textit{stride 2}) \vspace{1.0em} \\
                \bottomrule
            \end{tabular}
            \caption{
                VAE encoder and decoder architectures. %
                The model's inputs and outputs change based on the number of channels in the dataset, while the number of latent units the model has depends on the experiment hyper-parameters.
            }
            \label{fig:model-architecture}
        \end{table}

    \subsection{Optimiser And Batch Size}
    
        Models are trained using the Adam~\citep{kingma2014adam} optimiser with a learning rate of $10^{-3}$. A batch size of 256 is used in the case of the $\beta$-VAE~\citep{higgins2017beta}. Similarly, in the case of the weakly-supervised Ada-GVAE~\citep{locatello2020weakly}, 256 observation pairs are sampled per batch using the strategy from \Cref{appendix:sampling-adavae-pairs}.

    \subsection{Experiment Sweeps}

        Experiment plots and results are all produced from models trained over grid searches of hyper-parameters. Grid search values are given in \Cref{fig:experiment-hparams}. If values are not specified in the hyper-parameter sweep, then default values from the corresponding section of the experiment or supplementary material are used.

        \begin{table*}[b]
            \centering
            \footnotesize
            \begin{tabular}{ccl}
                \toprule
                    \textbf{Experiment} &
                    \textbf{Total} &
                    \textbf{Hyper-Parameters} \\
                \midrule \\[0em]
                    \makecell[c]{
                        5.3. Example Adversarial Dataset Results \\
                        (Figure 6)
                    }
                    & \makecell{
                        $8\times2\times2\times5$ \\
                        $= 160$ \\
                        $\times 1 \;\text{repeats}$ \\
                        $= 160$ \\
                        \\
                        $\times \sim 4 \text{h}$ \\ %
                        $\approx 640\text{h}$
                    }
                    & \makecell[l]{
                        $\begin{aligned}
                            \text{train steps}    & = 115200 \\
                            \text{beta}\;(\beta)  & \in \begin{aligned} \{
                                                            & 0.000316, 0.001, 0.00316, \\
                                                            & 0.01, 0.0316, 0.1, 0.316, 1.0
                                                        \} \end{aligned} \\
                            \text{framework}      & \in \{\beta\text{-VAE}, \text{Ada-GVAE}\} \\
                            \text{latents}\;(\sZ) & \in \{9, 25\} \\
                            \text{dataset}        & \in \begin{aligned} \{
                                                            & \text{dSprites}, \text{3D Shapes}, \text{Cars3D}, \\
                                                            & \text{Small NORB}, \text{XYSquares}
                                                        \} \end{aligned} \\
                        \end{aligned}$
                    } \vspace{1.0em} \\ \hline \\[0em]
                    \makecell[c]{
                        5.4. Example of Varying Levels of Overlap\\
                        (Figure 9)
                    }
                    & \makecell{
                        $2\times2\times8$ \\
                        $= 32$ \\
                        $\times 5 \;\text{repeats}$ \\
                        $= 160$  \\
                        \\
                        $\times \sim 2 \text{h}$ \\  %
                        $\approx 320\text{h}$
                    }
                    & \makecell[l]{
                        $\begin{aligned}
                            \text{train steps}    & = 57600 \\
                            \text{beta}\;(\beta)  & \in \{ 0.001, 0.00316 \} \\
                            \text{framework}      & \in \{\beta\text{-VAE}, \text{Ada-GVAE}\} \\
                            \text{latents}\;(\sZ) & = 9 \\
                            \text{dataset}        & = \text{XYSquares} \\
                            \text{grid spacing}   & \in \{ 8, 7, 6, 5, 4, 3, 2, 1 \}
                        \end{aligned}$
                    } \vspace{1.0em} \\ \hline \\[0em]
                    \makecell[c]{
                        6.1. Example Augmented Loss Experiments\\
                        (Figure 11)
                    }
                    & \makecell{
                        $2\times2\times2$ \\
                        $= 8$ \\
                        $\times 5 \;\text{repeats}$ \\
                        $= 40$ \\
                        \\
                        $\times \sim 2 \text{h}$ \\ %
                        $\approx 80\text{h}$
                    }
                    & \makecell[l]{
                        $\begin{aligned}
                            \text{train steps}    & = 57600 \\
                            \text{beta}\;(\beta)  & \in \{ 0.0001, 0.0316 \} \\
                            \text{framework}      & \in \{\beta\text{-VAE}, \text{Ada-GVAE}\} \\
                            \text{latents}\;(\sZ) & = 25 \\
                            \text{dataset}        & = \text{XYSquares} \\
                            \text{recon. loss}    & \in \{ \text{MSE}, \text{BoxBlurMSE} \} \\
                            \text{box blur radius}& = 31 \;(\text{63x63 in size}) \\
                            \text{box blur weight}& = 63^2 = 3969 \\
                        \end{aligned}$
                    } \vspace{1.0em} \\
                \bottomrule
            \end{tabular}
            \caption{
                Grid search hyper-parameters used for the different experiments throughout this paper. %
            }
            \label{fig:experiment-hparams}
        \end{table*}

    \subsection{Total Compute}

        We estimate that approximately $\sim 1040$ hours of compute across a computing cluster have been used to train the models needed to generate the plots and results presented throughout this paper.

        Due to the inherent high variance of unsupervised VAE results, multiple runs using the same hyper-parameters but different random seeds are needed for comparing frameworks \citep{locatello2018challenging}. This susceptibility of unsupervised methods to the starting random seed makes extended comparisons between frameworks prohibitive due to the computational cost.

\end{document}